\documentclass[runningheads]{llncs}
\usepackage[most]{tcolorbox}
\usepackage{tikz}
\usepackage{amsmath}
\usepackage{amssymb}
\usepackage{pifont}
\newcommand{\xmark}{\ding{55}}%
\usepackage{booktabs}
\usepackage{float} 
\usepackage{multirow}
\usepackage{graphicx}
\usepackage{subfig}
\usepackage{wrapfig}
\usepackage[normalem]{ulem}
\usepackage{hyperref}
\usepackage{orcidlink}

\usepackage{times}
\usepackage{latexsym}
\usepackage{array}
\useunder{\uline}{\ul}{}

\usepackage[draft,textsize=footnotesize,textwidth=15mm]{todonotes}

\newcommand{\hw}{HATE-WATCH}

\DeclareMathOperator*{\argmax}{arg\,max}
\begin{document}
\usetikzlibrary{arrows.meta}

\title{Cross-Platform Hate Speech Detection with Weakly Supervised Causal Disentanglement}

%cross-platform Hate Speech Detection with Weakly Supervised Causal Disentanglement
%
\titlerunning{Cross-Platform Hate-Speech Detection HATE-WATCH}
% If the paper title is too long for the running head, you can set
% an abbreviated paper title here
%
% \author{First Author\inst{1}\orcidID{0000-1111-2222-3333} \and
% Second Author\inst{2,3}\orcidID{1111-2222-3333-4444} \and
% Third Author\inst{3}\orcidID{2222--3333-4444-5555}}
% %
% \authorrunning{F. Author et al.}
% % First names are abbreviated in the running head.
% % If there are more than two authors, 'et al.' is used.
% %
% \institute{Princeton University, Princeton NJ 08544, USA \and
% Springer Heidelberg, Tiergartenstr. 17, 69121 Heidelberg, Germany
% \email{lncs@springer.com}\\
% \url{http://www.springer.com/gp/computer-science/lncs} \and
% ABC Institute, Rupert-Karls-University Heidelberg, Heidelberg, Germany\\
% \email{\{abc,lncs\}@uni-heidelberg.de}}
%
\titlerunning{HATE-WATCH}
\authorrunning{Sheth et al.}
\author{Paras Sheth\inst{1}\orcidlink{0000-0002-6186-6946} \and
Tharindu Kumarage\thanks{Both authors contributed equally.}\inst{1}\orcidlink{0000-0002-9148-0710} \and
Raha Moraffah$^\star$\inst{1}\orcidlink{0000-0002-6891-2925} \and
Aman Chadha\thanks{Work does not relate to the position at Amazon}\inst{2,3}\orcidlink{0000-0001-6621-9003} \and
Huan Liu\inst{1}\orcidlink{0000-0002-3264-7904}}

\institute{Arizona State University, Tempe, AZ, USA \and
Stanford University, Stanford, CA, USA \and Amazon Alexa AI, Sunnyvale, CA, USA  \\
\email{\{psheth5,kskumara,rmoraffa,huanliu\}@asu.edu}\\
\email{hi@aman.ai}}

\maketitle              % typeset the header of the contribution
\begin{abstract}
   %How online social media is misused for hate. 
   %How causality helps and whats the less label issue?
   %How can we tackle that issue?
   Content moderation faces a challenging task as social media's ability to spread hate speech contrasts with its role in promoting global connectivity. With rapidly evolving slang and hate speech, the adaptability of conventional deep learning to the fluid landscape of online dialogue remains limited. In response, causality-inspired disentanglement has shown promise by segregating platform-specific peculiarities from universal hate indicators. However, its dependency on available ground-truth target labels for discerning these nuances faces practical hurdles with the incessant evolution of platforms and the mutable nature of hate speech. Using confidence-based reweighting and contrastive regularization, this study presents {\hw}, a novel framework of weakly supervised causal disentanglement that circumvents the need for explicit target labeling and effectively disentangles input features into invariant representations of hate. Empirical validation across platforms (two with target labels and two without) positions {\hw} as a novel  method in cross-platform hate speech detection  with superior performance. {\hw} advances scalable content moderation techniques towards developing safer online communities.
\end{abstract}
\section{Introduction}
\textbf{Warning:} \textit{This paper contains contents that may be offensive or upsetting.}
% \\
%Hate speech's definition and its evolution over time and why its spread needs to be brought to a halt to create safer communities (Try to talk about how hate speech's existence is a problem for safe communities).
% The proliferation of hate speech on social media not only undermines the principles of healthy discourse but also poses a tangible threat to societal safety and cohesion. Acts of violence and prejudice motivated by hate speech on the internet have spread from virtual to physical realms, highlighting the connection between hate speech online and its effects in the real world. Events like the 2019 killings at the Christchurch mosque in New Zealand~\cite{heie2020cyclical,macklin2019christchurch}, for example, show how the assailant was radicalized by extremist information on the internet which led to the deadly consequences due to unrestrained hate speech. In a similar vein, the rise in hate crimes against Asian populations was seen throughout the COVID-19 outbreak~\cite{cheah2020covid,gover2020anti}. The hate crimes were exacerbated by xenophobic remarks on social media as another evidence of how virtual hatred can materialize into actual violence and societal divisions. These examples showcase the need for effective moderation strategies that can adapt to the dynamic nature of online discourse and aid in mitigating the spread of hate speech.

Hate speech on social media not only disrupts healthy dialogue but also threatens societal harmony and safety. It has been linked to real-world violence, as seen in events like the 2019 Christchurch mosque shootings in New Zealand, where the perpetrator was influenced by online extremist content~\cite{macklin2019christchurch}. Similarly, the COVID-19 pandemic saw a spike in hate crimes against Asians, fueled by xenophobic comments online~\cite{cheah2020covid}. These instances underscore the urgent need for advanced moderation tools to curb the spread of online hate and prevent its escalation into physical aggression and division.

%Existing SOTA's how they function and why they fail to generalize. Then talk about existing SOTA that can generalize and what are their shortcomings.

The identification of hate speech on social media poses considerable hurdles because of the rapid expansion of these platforms~\cite{jeong2023exploring} and the emergence of distinct community standards. Lacking platform-specific labeled data, new platforms frequently rely on models that were trained using data from existing platforms~\cite{das2020hate} or by directly leveraging large language models (LLMs)~\cite{kikkisetti2024using}. Utilizing these solutions has pitfalls as LLMs might not be able to determine hate due to the social-context demands of the task~\cite{yuan2024revisiting}. On the other hand, the hate speech models trained on one platform may not generalize to other platforms due to various factors, including; (1) the variance in regulatory policies, from stringent content controls on platforms like FoxNews~\cite{caplan2018dead} to more lenient policies on sites like GAB~\cite{buckley2022censorship}, and (2) hate speech's context-dependent characteristics. For example, the COVID-19 pandemic's increase in hate speech against Asians brought to light the limitations of pre-pandemic models, which lacked the contextually relevant training data to effectively identify and mitigate such emerging forms of hate~\cite{he2021racism}. These complexities illustrate the challenges in creating adaptable and accurate hate speech detection models across varying social media environments.

Efforts to enhance cross-platform model performance have explored linguistic cues~\cite{ramponi-tonelli-2022-features} and context-aware fine-tuning~\cite{caselli2020hatebert}. Yet, these methods often fall short, either forming spurious correlations or embedding platform-specific biases that hinder generalization. Recognizing these limitations, recent causality-based approaches have emerged as promising alternatives, leveraging causal cues~\cite{sheth2023peace} or graphs~\cite{sheth2024causality} to guide models in learning platform-invariant hate representations. While these causality-based methods address several of the aforementioned challenges, they introduce significant demands for data annotation and comprehensive auxiliary data, such as the precise target of hatred, raising concerns regarding feasibility, privacy, and ethical implications.

%Intuitively what can one do to address these problems. Based on that introduce causality and technical aspects of the solution.
In response to these challenges and building upon the foundation laid by causality-based approaches, we propose a novel \textbf{W}eakly supervised c\textbf{A}usality-aware disen\textbf{T}anglement framework for \textbf{C}ross-platform \textbf{H}ate speech detection named {\hw}\footnote{The code can be accessed from \url{https://anonymous.4open.science/r/HATE-WATCH-406C/HATE_WATCH.ipynb}}, that synthesizes weak supervision with contrastive learning to minimize the dependency on extensively labeled target data. Inspired by advancements in fine-tuning language models (LMs) with weak supervision~\cite{yu2020fine}, {\hw} aims to mitigate the challenges of noisy labels and sparse data, significantly enhancing the model's adaptability across various social media platforms. By leveraging contrastive learning {\hw} facilitates distinguishing between different target representations. Furthermore, with confidence-based sample reweighting methods {\hw} suppresses label noise propagation during training ensuring effective model learning. This approach not only simplifies the demands for auxiliary data but also leverages the inherent variability and noise within weakly labeled datasets as an asset, thereby improving the robustness and generalization of hate speech detection.

{\hw} represents a significant advancement in cross-platform hate speech detection research by offering a scalable and equitable solution that addresses the practical and ethical concerns associated with current methodologies. Our contributions are summarized as follows:
\begin{itemize}
    \item We introduce a causality-based detection framework {\hw} enhanced by weak supervision and contrastive learning, significantly reducing the reliance on labeled data for platform-dependent features.
    \item {\hw} demonstrates superior adaptability and generalization in hate speech detection across four social media environments (two with target labels and two without), validated through comprehensive experiments.
    % \item Our methodology offers both ethical and practical benefits for scalable content moderation, presenting a novel approach to combating online hate speech effectively.
\end{itemize}
%My contributions
\section{Related Work}

% \hlnote{How about disentanglement work?}

% Recent advancements in hate-speech detection focus on enhancing generalization across diverse social media platforms and target demographics, crucial due to the dynamic and varying nature of online hate speech. Generalization challenges stem from the wide range of hate-speech manifestations, with models often struggling to detect new topics~\cite{corazza2019cross,ali2022hate} and target groups~\cite{pamungkas2021joint,del2017hate}.
Recent strides in hate speech detection emphasize generalizability across varying social media landscapes and demographics, addressing the challenge posed by the dynamic nature of online hate speech. The diversity in hate speech manifestation complicates detection of emerging topics and groups~\cite{pamungkas2021joint}.

\noindent \textbf{Enhancing Cross-platform Hate Speech Detection via Representation Disentanglement:} Addressing the scarcity of widely available data, recent efforts have harnessed auxiliary features such as user attributes~\cite{del2023socialhaterbert} and annotator demographics~\cite{yin2022annobert}, alongside LMs including Llama 2, GPT-3.5~\cite{kumarage2024harnessing}, BERT, and RoBERTA~\cite{caselli2020hatebert,mathew2021hatexplain} for cross-platform hate speech detection. However, the absence of explicit training objectives for generalizability in these detectors could lead to errors in cross-platform scenarios~\cite{sheth2023peace}. Recent studies aim to enhance generalizability through explicit modeling of hate representations, using causal cues like aggression and sentiment~\cite{sheth2023peace}

Recent advances in representation disentanglement, encoding distinct aspects of data~\cite{bengio2013representation}, show significant promise in generalizing hate speech detection across modalities by separating hate-target entities~\cite{lee2021disentangling,yang2022multimodal} and distinguishing genuine from spurious content elements~\cite{ramponi2022features}. Notably, causality-guided disentanglement~\cite{sheth2024causality}, inspired by the platform-specific nature of hate targets~\cite{mansur2023twitter,carvalho2023expression}, uses invariant features for state-of-the-art generalization. This approach demonstrated that by removing platform-dependent features and leveraging the invariant features one can achieve state-of-the-art generalization, though it often presupposes available hate-target annotations. Therefore, our study aims to explore a more pragmatic approach, where ground-truth target labels are absent.

% demonstrated that by excluding hate target representations from the LM-based detector's representation, the models become invariant to platform changes, thereby achieving state-of-the-art performance in generalized hate speech detection~\cite{sheth2024causality}. However, a significant challenge in this area of research is the assumption that hate-target annotations are readily available, which is often not the case in social media platforms. Therefore, our study aims to explore a more pragmatic approach by working in settings where ground-truth human-annotated target labels are absent, yet we manage to obtain some noisy labels from a weak target classifier.

\noindent \textbf{Weakly Supervised Learning}  In the face of inconsistent data and varying annotations, this strategy refines noisy labels, which is advantageous for hate speech models~\cite{jin2023towards}. The rise of pre-trained LMs has facilitated the emergence of weakly supervised methods, such as Xclass~\cite{wang2020x} and LOTClass~\cite{meng2020text}, which leverage contextual information from label names and keywords for denoising and learning enhancement. Recent studies have demonstrated the effectiveness of these methods in achieving cross-dataset generalization in hate speech detection~\cite{jin2023towards}. Current developments demonstrate that by leveraging contrastive self-training, a technique that harnesses the denoising capabilities of pre-trained LMs widens their applicability to a broader array of tasks~\cite{yu2020fine}. In line with these advancements, our research adopts the contrastive self-training method to learn cross-platform hate speech representations with weak target labels.

\section{Methodology}
\subsection{Preliminaries}

\begin{wrapfigure}{r}{0.5\textwidth}
\centering
\footnotesize
\begin{tikzpicture}
\begin{scope}[every node/.style={circle,thick,draw,minimum size=1cm}]
\node (X) at (-2,0.5) {$X$};
\node [dashed] (P) at (-2,3) {$P_{l}$};
\node (Xw) at (-4,1.5) {$X_{t}$};
\node (Xc) at (0,2.5) {$X_{c}$};
\node (Y) at (0,0.5) {$Y$};
\end{scope}

\begin{scope}[>={Stealth[black]},
              every edge/.style={draw=black,very thick}]
    \path [->] (P) edge node {} (Xw);
    \path [->] (Xw) edge node {} (X);
    \path [->] (Xc) edge node {} (X);
    \path [->] (Xc) edge node {} (Y);
\end{scope}
\end{tikzpicture}
\caption{The causal graph illustrates the hate speech detection mechanism, where $X_c$ denotes the causal factors predicting hate speech, $Y$ the hate label, $X_w$ the target, $X$ the input, and $P_l$ the latent platform variable affecting the target.}
\label{cg}
\end{wrapfigure}
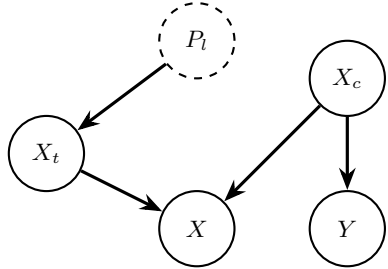

Separating platform-specific from invariant features is crucial for identifying hate speech across platforms. Causality, a tool proven to boost a model's generalization~\cite{buhlmann2020invariance}, can underpin this process, as shown in Figure~\ref{cg}. Here, the input data can be decomposed into (1) a causal component, constant across platforms (e.g., aggression) and (2) a platform-specific component, unique to each platform (e.g., platform policies).

Distinguishing between these factors is pivotal, yet quantifying platform policies is challenging. Previous work suggests using hate targets as proxies for platform-specific aspects~\cite{sheth2024causality}. However, acquiring auxiliary and ground-truth labels is difficult, especially on data-scarce platforms. Additionally, trying to disentangle in an unsupervised manner yields incorrect disentanglement rendering the model ineffective~\cite{locatello2020sober}. 

To overcome these barriers, we introduce {\hw}, which bypasses the need for target labels by adopting a weakly supervised, contrastive learning method with confidence-based reweighting for precise disentanglement. Given a source corpus $D_{source}$ consisting of textual inputs $X = \{x_{1}, x_{2}, \ldots, x_{n}\}$ and corresponding hate speech labels $Y = \{y_{1}, y_{2}, \ldots, y_{n}\}$, {\hw} aims to establish a mapping $F: X \rightarrow Y$ that pinpoints the causal attributes of hate speech. It employs contrastive learning to maintain distinct target representations and integrates confidence-based reweighting to reduce noise impact, enhancing model training's effectiveness.

\subsection{Disentangling Causal and Target representations}
% In the landscape of online discourse, 
The manifestation of hate speech often intertwines with the targets it seeks to disparage. Even though the precise targets could fluctuate significantly between platforms—a reflection of the platform-dependent nature of hate speech—the fundamental organization and severity of the offensive material frequently show striking consistency. For instance, foul language used to disparage one group can frequently be applied with little modification to another target. This finding forms the basis of our methodology in {\hw}, where we propose that the essential elements of hate—its causal features—can be successfully extracted from the text even in the lack of clear hate target labeling. The goal of {\hw} is to capture the substance of hate speech by abstracting from the changeable, platform-specific targets. This allows to identify and analyze hateful content with an intuitive knowledge that goes beyond specific targets.

% The {\hw} architecture consists of three main components: an encoder, a disentanglement mechanism, and a decoder. These components are designed to adhere to the typical VAE structure, but are specifically tuned for weak supervision. The encoder, denoted as $q_{\phi}$, is a language model e.g., RoBERTa~\cite{liu2019roberta} which processes an input text $x$ to produce an embedding $z$ represented as,
% $z = q_{\phi}\left(\gamma(x)\right)$,
% where $z \in \mathbb{R}^{s_{l} \times h_{d}}$ is the resultant embedding from the language model, $\gamma(x)$ denotes the tokenization function, $s_{l}$ is the sequence length, and $h_{d}$ signifies the dimension of the embedding. The embedding of the [CLS] token, placed at the initial position and represented by $z^{[CLS]} \in \mathbb{R}^{h_{d}}$, serves as the representative input for $x$.
\begin{figure}[t]
\centering
\includegraphics[width=0.95\columnwidth]{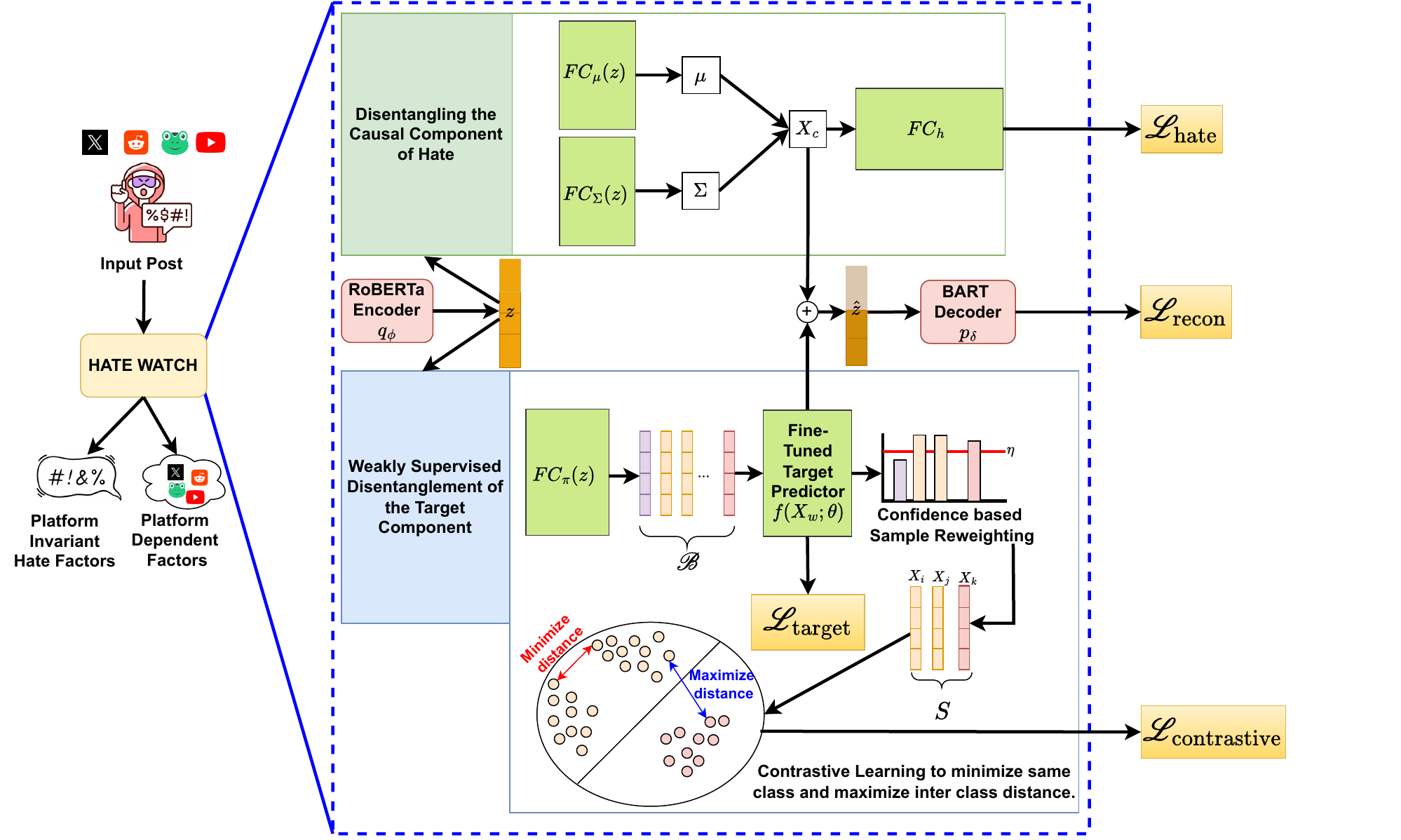}
% \acil{Let's use the vectorized/PDF version of the above diagram; this one is a JPG hence rasterized (zoom-in etc. is limited for readability with rasterized diagrams).}
     \caption{The {\hw} architecture processes input $X$ via a RoBERTa to get initial representation $z$. This $z$ undergoes disentanglement into a causal component to identify invariant hate factors $X_c$, and a weakly supervised target component $X_w$ without true target labels. Both components' outputs, $X_c$ and $X_w$, are merged to form reconstructed embedding $\hat{z}$, which is decoded by BART to produce reconstructed input $\hat{X}$.}
\label{fig:architecture}
\end{figure}
{\hw} utilizes a three-component architecture—encoder, disentanglement mechanism, and decoder—mirroring a VAE structure optimized for weak supervision. The encoder $q_{\phi}$, a language model like RoBERTa~\cite{liu2019roberta}, transforms input text $x$ into an embedding $z$ via $z = q_{\phi}\left(\gamma(x)\right)$, where $z \in \mathbb{R}^{s_{l} \times h_{d}}$ is the embedding output, $\gamma(x)$ is the tokenization of $x$, $s_{l}$ represents sequence length, and $h_{d}$ the embedding dimension. The \texttt{[CLS]} token embedding $z^{[CLS]} \in \mathbb{R}^{h_{d}}$, indicating the input's overall representation, exemplifies this process. The overall framework of HATE-WATCH is visualized in Figure.~\ref{fig:architecture}.

\textbf{Disentangling the Causal Component}\quad {\hw} aims to separate the embedding $z$ in order to extract the subset of causal features $X_c$ using its VAE design. Disentanglement is a procedure that seeks to derive a factorized representation by aligning individual latent variables with distinct explanatory factors of the data. The separation process is facilitated by the typical Gaussian prior applied to the latent space inside the VAE framework, as well as the posterior approximation achieved by a parameterized neural network. In this study, two feedforward neural networks, namely $FC_{\mu}$ and $FC_{\Sigma}$, are utilized to map the variable $z$ onto the parameters of a Gaussian distribution. This mapping is designed to capture the inherent causal relationship present in hate speech. The distribution is used to create the latent causal feature set $X_{c} \in \mathbb{R}^{h_{causal}}$, employing the re-parameterization trick~\cite{kingma2014semi} as follows:
\begin{equation}
X_{c}= Enc_{1}(\mu_z,\Sigma_z) = \mu_{z} + \Sigma_{z} \odot \epsilon \sim \mathcal{N}(\mathbf{0}, \mathbf{I}),
\end{equation}
where $\mu_{z} = FC_{\mu}(z)$, $\Sigma_{z} = FC_{\Sigma}(z)$, $\epsilon$ is a random noise vector sampled from $\mathcal{N}(\mathbf{0}, \mathbf{I})$, representing the standard normal distribution, and  $\mathbf{I}$ is the unit variance.

\textbf{Disentangling the Target Component}\quad
In addition to the causal features captured by the embedding $z$, our model must contend with platform-dependent features that vary across different social media environments, denoted by $X_{w}$. Given that these factors are not easily quantifiable (e.g., policies and regulations of a platform) we leverage a proxy, i.e., the target of hate based on the observation that the distribution of the targets is platform dependent~\cite{doring2019male,waseem2016you}. 

Unlike the invariant causal features, these platform-dependent attributes are inherently discrete, with hate speech targeting specific categories such as Race, Religion, Gender, among others~\cite{bourgeade2023did}. Given the discrete nature of targets and the absence of ground truth labels in various real-world scenarios, we employ a contrastive learning framework along with confidence based weighted resampling, inspired by~\cite{yu2020fine} to learn a distinct representation for each target class.

\subsubsection{Contrastive Learning Framework}
The essence of our approach is to enforce a structure within the latent space that brings closer the representations of samples with similar targets while distancing those with different ones. To implement this, we utilize a feed-forward neural network $FC_{\pi}$ to map the embedding $z$ onto a latent space $X_{w}$ where, $X_w = FC_{\pi}(z)$,
and $X_{w} \in \mathbb{R}^{h_{disc}}$ holds the latent representation for the target classes. This representation is further passed through a target classifier $f(X_{w}; \theta)$ which is a finetuned feed-forward network, to obtain the soft-labels $\tilde{y}$.

Given the soft pseudo-target labels $\tilde{y}$ generated by the model, we define a contrastive loss that encourages a compact cluster formation for each target class. Specifically, we first select high-confidence samples $\mathcal{S}$ from $\mathcal{B}$ as $\mathcal{S} = \{ X_{w} \in \mathcal{B} \mid w(x) \geq \eta \}$
where $\eta$ is a predefined threshold. Then, we define the similarity between each pair $X_{{w}_{i}}, X_{{w}_{j}} \in \mathcal{S}$ as follows:
\begin{equation}
W_{ij} = \begin{cases} 
1, & \text{if } \argmax_{k \in \mathcal{Y}}[\tilde{y}_i]_k = \argmax_{k \in \mathcal{Y}}[\tilde{y}_j]_k, \\
0, & \text{otherwise},
\end{cases}
\end{equation}
where $\tilde{y}_i, \tilde{y}_j$ are the soft pseudo-labels obtained from for $X_{{w}_{i}}, X_{{w}_{j}}$, respectively.
Next, we define the contrastive regularizer as follows: 
% \begin{equation}
% \mathcal{L}_{contrastive} = \sum_{(X_{{w}_{i}}, X_{{w}_{j}}) \in \mathcal{S} \times \mathcal{S}} \ell(X_{{w}_{i}}, X_{{w}_{j}}, W_{ij}),
% \end{equation}
% where
% \begin{equation}
% \ell = W_{ij}d_{ij}^2 + (1 - W_{ij})\max(0, \beta - d_{ij})^2.
% \label{contr}
% \end{equation}
% The contrastive loss~\cite{chopra2005learning,taigman2014deepface} in this case is represented by $\ell(\cdot,\cdot)$, the Euclidean distance between $X_{{w}_{i}}, X_{{w}_{j}}$ is $d_{ij}$, and a pre-defined margin is $\beta$.
\begin{equation}
\begin{aligned}
\mathcal{L}_{contrastive} &= \sum_{(X_{{w}_{i}}, X_{{w}_{j}}) \in \mathcal{S} \times \mathcal{S}} \left[ W_{ij}d_{ij}^2 + (1 - W_{ij})\max(0, \beta - d_{ij})^2 \right], \\
\ell &= W_{ij}d_{ij}^2 + (1 - W_{ij})\max(0, \beta - d_{ij})^2,
\end{aligned}
\label{contr}
\end{equation}
where the contrastive loss \(\mathcal{L}_{contrastive}\) utilizes \(\ell(\cdot,\cdot)\) to measure the Euclidean distance (\(d_{ij}\)) between \(X_{{w}_{i}}, X_{{w}_{j}}\), and \(\beta\) denotes a pre-defined margin. The function \(\ell\) is emphasized for its use in calculating the individual contributions to the overall contrastive loss, as detailed in works such as~\cite{chopra2005learning,taigman2014deepface}.

Eq. \ref{contr} penalizes the distance between samples from the same class, i.e., $W_{ij} = 1$; for samples from different classes, the contrastive loss is considerable if their distance is smaller. The regularizer maintains at least $\gamma$ distance between dissimilar samples and forces similar samples to be close together in this fashion.

\subsubsection{Confidence-Based Sample Reweighting}
The contrastive loss facilitates learning distinct representations for each target class, however in the absence of ground truth labels it may result in label-noise propagation~\cite{ren2020denoising}. To mitigate the propagation of label noise due to the weakly supervised setting, we incorporate a confidence-based sample reweighting mechanism. Each sample $i$ is assigned a weight based on the entropy of its predicted probability distribution $\tilde{y}_i$, reflecting the confidence of the model in its prediction. The weights are calculated as $w_i = 1 - \frac{H(\tilde{y}_i)}{\log(C)},$ where $H(\tilde{y}_i)$ denotes the entropy of the predicted probabilities for sample $i$, and $C$ is the number of target classes. 

During contrastive self-training, the sample reweighting strategy encourages high confidence samples. This approach, however, depends on incorrectly classified samples being considered low confidence, which may not be the case unless we stop overly optimistic forecasts. To promote smoothness over forecasts, we utilize a confidence based regularizer as,
\begin{equation}
\mathcal{L}_{conf} = \frac{1}{|\mathcal{S}|} \sum_{X_{w} \in \mathcal{S}} \mathcal{D}_{KL} (u || f(X_{w}; \theta)),
\end{equation}
where \( \mathcal{D}_{KL} \) is the KL-divergence and \( u_i = \frac{1}{C} \) for \( i = 1, 2, \ldots, C \), and $C$ is the total target classes. Such term constitutes a regularization to prevent over-confident predictions and leads to better generalization~\cite{pereyra2017regularizing}. We thus define the loss function as,
\begin{equation}
\mathcal{L}_{target} (\theta, \tilde{y}) = \frac{1}{|\mathcal{S}|} \sum_{X_{w} \in \mathcal{S}} w(X_{w}) D_{KL} (\tilde{y} || f(Z_{\pi}; \theta)),
\end{equation}
where
% \begin{equation}
% D_{KL} (P || Q) = \sum_k p_k \log \frac{p_k}{q_k}
% \end{equation}
$D_{KL}$ is the Kullback-Leibler (KL) divergence~\cite{kullback1997information}.

By integrating the contrastive strategy with confidence-based sample reweighting, our model {\hw} is capable of learning to disentangle target representations that are robust to variations in data distribution and labeling noise.
\subsubsection{Reconstructing the Input from the Disentangled components}
To facilitate the training, we aim to reconstruct the input from the obtained components $X_c$ and $X_w$. To do so, we first concatenate them together as $[X_c \,|\, X_w]$ where $[ \,|\, ]$ is the concatenation operation. Then we pass it through another feed-forward neural network, namely, $FC_{\hat{z}}$. The obtained representation after concatenation is given by $\hat{z} = FC_{\hat{z}}([X_c \,|\, X_w])$. To recreate the input, we feed $\hat{z}$ through a LM decoder $p(x|\hat{z})$. We utilize BART-base decoder~\cite{lewis2019bart} as the LM-decoder, as it shares the vocabulary with RoBERT~\cite{liu2019roberta} implementations and is proved powerful in many generative tasks. The obtained tokens are given as 
$\hat{x} = LMHead(p_{\delta}(\hat{z}))$,
where \(LMHead\) is a feed-forward layer to map the decoder \(p_{\delta}\) embeddings into tokens. The reconstructed loss is computed between the input token ids and the reconstructed token ids and is formulated as follows:
\begin{equation}
L_{\text{recon}}(y(x), \hat{x}) = - \sum_{i=1}^{s_{l}} y(x) \log(\hat{x}_i),
\end{equation}
where $s_{l}$ is the sequence length. We also apply KL-divergence losses for both latent spaces to make sure that the posteriors of the disentangled latent spaces are near to their prior distribution. The disentanglement module's Evidence Lower BOund (ELBO) is expressed as follows:
\begin{equation}
    \mathcal{L}_{VAE} = \mathcal{L}_{recon} + \alpha_{t}*\operatorname{L}_{\mathbb{D}_{target}} + \alpha_{c}*\operatorname{L}_{\mathbb{D}_{causal}},
\end{equation}
where $\alpha_{t}$ represents the coefficient that controls the contribution of the KL loss for the target, and $\alpha_{c}$ represents the coefficient that controls the contribution of the causal KL loss. To facilitate learning in absence of ground-truth target labels, and by leveraging confidence based denoising and contrastive learning inspired by~\cite{yu2020fine} $\operatorname{L}_{\mathbb{D}_{target}}$ and $\operatorname{L}_{\mathbb{D}_{causal}}$ are given by,
% \begin{equation}
% \begin{aligned}
% \operatorname{L}_{\mathbb{D}_{target}} &= \mathcal{L}_{target} + \delta_{cont} * (\mathcal{L}_{contrastive}) + \delta_{conf} * (\mathcal{L}_{conf})
% \end{aligned}
% \end{equation}
% Similarly, $\operatorname{L}_{\mathbb{D}_{causal}}$ is given by, 
% \begin{equation}
% \begin{aligned}
% \operatorname{L}_{\mathbb{D}_{causal}} &= D_{\mathrm{KL}}\left(Enc_{1}(X_{c} \mid X) \| p(X_{c})\right)
% \end{aligned}
% \end{equation}
\begin{equation}
\begin{aligned}
\operatorname{L}_{\mathbb{D}_{target}} &= \mathcal{L}_{target} + \delta_{cont} \cdot \mathcal{L}_{contrastive} + \delta_{conf} \cdot \mathcal{L}_{conf}, \\
\operatorname{L}_{\mathbb{D}_{causal}} &= D_{\mathrm{KL}}\left(Enc_{1}(X_{c} \mid X) \| p(X_{c})\right),
\end{aligned}
\end{equation}

\subsection{Model Training}
The disentangled latent causal representation $X_c$ is used to calculate the classification probability for hate speech detection, given by $\hat{y}_i=\operatorname{Softmax}(FC_{h}(X_{c}))$
where $FC_{h}$ represents a fully connected layer for hate classification. The total loss $\mathcal{L}$ is then calculated using the conventional cross-entropy method:
\begin{equation}
\mathcal{L}_{hate}=-\frac{1}{N} \sum_{i=1}^{|D_{source}|} y_i \log \hat{y}_i    
\end{equation}
where $D_{source}$ represents the source domain data, $y_i$ represents the true hate label, and $\hat{y}_i$ denotes the predicted hate labels.  Lastly, we integrate every suggested module and train using a multi-task learning approach:
\begin{equation}
\mathcal{L}=\mathcal{L}_{hate}+ \mathcal{L}_{VAE}
\end{equation}

\begin{table}[t]
\centering
\footnotesize
\setlength{\tabcolsep}{4pt} % Adjust the padding between columns
\begin{tabular}{lcccc}
\toprule
\textbf{Dataset} & \textbf{\# Posts} & \textbf{Hateful Posts} & \textbf{Hate \%} & \textbf{Target?}\\ 
\midrule
\textbf{GAB}~\cite{mathew2021hatexplain}      & 11,093                & 8,379                  & 75.5          & \checkmark   \\ 
\textbf{Reddit}~\cite{kurrek2020towards}   & 39,811                & 15,388                 & 38.6       & \xmark      \\ 
\textbf{X}~\cite{davidson2017automated}  & 24,802                 & 9,118                  &   36.7    & \xmark       \\ 
\textbf{YouTube}~\cite{salminen2018anatomy}  & 1,026                 & 642                    & 62.5      & \checkmark       \\ 
\bottomrule
\end{tabular}
\caption{Dataset statistics with percentage of hateful comments or posts.}
\label{datasets}
\end{table}

\section{Experiments}
In this section, we conduct a series of experiments aimed at verifying the capability of {\hw} in acquiring generalizable representations to identify hate speech through causality-aware disentanglement without reliance on target labels. We use benchmark datasets from multiple platforms to guarantee a comprehensive analysis. We aim to answer the following research questions:
\begin{itemize}
    \item \textbf{RQ.1} Is it possible to effectively disentangle the causal and platform dependent factors even in the absence of ground truth target labels?
    \item \textbf{RQ.2} How does forgoing target labels affect disentanglement effectiveness compared to using labels derived from LLMs?
    \item \textbf{RQ.3} Does weakly-supervised disentanglement learn invariant relationships?
\end{itemize}

% \begin{table}[H]
% \centering
% \footnotesize
% \begin{tabular}{ccccc}
% \toprule
% \textbf{Datasets} & \textbf{No. of Posts} & \textbf{Hateful Posts} & \textbf{Hate \%} & \textbf{Target Label}\\ 
% \midrule
% \textbf{GAB}~\cite{mathew2021hatexplain}      & 11,093                & 8,379                  & 75.5          & \cmark   \\ 
% % \hline
% \textbf{Reddit}~\cite{kurrek2020towards}   & 39,811                & 15,388                 & 38.6       & \xmark      \\ 
% % \hline
% \textbf{X}~\cite{davidson2017automated}  & 24,802                 & 9,118                  &   36.7    & \xmark       \\ 
% % \hline
% \textbf{YouTube}~\cite{salminen2018anatomy}  & 1,026                 & 642                    & 62.5      & \cmark       \\ \bottomrule
% \end{tabular}
% \vspace{1mm}
% \caption{Dataset statistics with corresponding platforms and percentage of hateful comments or posts.}
% \label{datasets}
% \end{table}

% We perform binary classification of hate speech detection on widely used benchmark hate datasets. Since we aim to verify cross-platform generalization, we use data from four different platforms for cross-platform evaluation: GAB, Reddit, X, and YouTube. All datasets are in the English language. GAB~\cite{mathew2021hatexplain} is a collection of annotated posts from the GAB website. It consists of binary labels indicating whether a post is hateful or not. These instances are annotated with corresponding explanations, where crowd workers justify why the given post, or content is considered hateful. Reddit~\cite{kurrek2020towards}
\subsection{Dataset and Evaluation Metrics}
To identify hate speech on English-language benchmark datasets from GAB, Reddit, X, and YouTube, binary classification is used. GAB, which is annotated for hatefulness and is obtained from the GAB website~\cite{mathew2021hatexplain}. The YouTube dataset, which has been described in detail by~\cite{salminen2018anatomy}, has comments that use offensive language. Hate and target labels are provided by both datasets.

Four ordinal labels are used in Reddit's hatefulness-categorized dataset~\cite{kurrek2020towards}. We classify any denigrating information as hateful for consistency's sake. In a similar vein, content is classified as Hate, Offensive, or Neither in X's dataset that was generated from tweets~\cite{davidson2017automated}, with the first two being deemed hateful. X and Reddit are strictly binaryized into hate and non-hate similar to the other platforms. Unlike GAB and YouTube, X and Reddit do not have any hate target labels. These datasets are compiled in Table~\ref{datasets}, and their evaluation is based on the macro F1-measure.

\subsection{Experimental Settings}
\noindent{\textbf{Implementation Details}}\quad
We trained our framework using RoBERTa-base for the LM-encoder and BART-base for the LM-decoder with the Huggingface Transformers library. We optimized the model with cross-entropy loss and AdamW, using a learning rate of 0.0001, dropout rate of 0.2, and parameters $\alpha_{t}$ as 0.05, $\alpha_{c}$ as 0.05, $\delta_{cont}$ as 0.001, $\delta_{conf}$ as 0.001, $\eta$ as 0.95 and $\beta$ as 2. Training was conducted on an NVIDIA GeForce RTX 3090 GPU with 24 GB VRAM, with early-stopping.

\noindent{\textbf{Baselines}}\quad In our analysis, the {\hw} framework is evaluated against leading methods across three categories: fine-tuned LMs, causality-aware techniques, and weakly supervised approaches, to benchmark its effectiveness in cross-platform hate speech detection. 
% Summaries of these baselines are as follows:

\noindent \textbf{Fine-Tuned LMs:} \textbf{HateXplain}~\cite{mathew2021hatexplain} and \textbf{HateBERT}~\cite{caselli2020hatebert} are optimized for hate speech identification, the former focusing on a trinary classification with annotated justifications using X and GAB data, and the latter finetuned on 1.5 million Reddit posts for training a BERT-base model.
    
\noindent \textbf{Causality-Aware Techniques:} Utilizing sentiment and aggression features (\textbf{PEACE}~\cite{sheth2023peace}) or auxiliary information like target labels (\textbf{CATCH}~\cite{sheth2024causality}), these methods aim to improve generalization for hate speech from a causal lens.
    
\noindent \textbf{Weakly Supervised Approaches:} \textbf{XClass}~\cite{wang2020x} and \textbf{LOTClass}~\cite{meng2020text} employ contextual cues and semantic relationships for noise reduction and category prediction, enhancing model training in the absence of explicit labels.

\addtolength{\tabcolsep}{1.5pt}
\begin{table}[t]
\resizebox{\columnwidth}{!}{%
\begin{tabular}{cccccccccc}
% \hline
\toprule
\multicolumn{1}{c}{\multirow{2}{*}{\textbf{Dataset Type}}}                                                        & \multicolumn{1}{c}{\multirow{2}{*}{\textbf{Source}}}  & \multirow{2}{*}{\textbf{Target}} & \multicolumn{7}{c}{\textbf{Models}}                                                                                                                                                                                                                          \\ \cmidrule{4-10} 
\multicolumn{1}{c}{}                                                                                              & \multicolumn{1}{c}{}                                  &                                  & \multicolumn{1}{c}{\textbf{HateXplain}} & \multicolumn{1}{c}{\textbf{HateBERT}} & \multicolumn{1}{c}{\textbf{PEACE}} & \multicolumn{1}{c}{\textbf{CATCH}} & \multicolumn{1}{c}{\textbf{XClass}} & \multicolumn{1}{c}{\textbf{LOTClass}} & \textbf{OURS} \\ 
% \hline
\midrule
\multicolumn{1}{c}{\multirow{7}{*}{\textbf{\begin{tabular}[c]{@{}c@{}}With\\ Target \\ Labels\end{tabular}}}}     & \multicolumn{1}{c}{\multirow{4}{*}{\textbf{GAB}}}     & \textbf{GAB}                     & \multicolumn{1}{c}{{\ul 0.87}}          & \multicolumn{1}{c}{\textbf{0.89}}     & \multicolumn{1}{c}{0.76}           & \multicolumn{1}{c}{0.82}           & \multicolumn{1}{c}{0.79}            & \multicolumn{1}{c}{0.77}              & 0.81          \\ % \cmidrule{3-10} 
\multicolumn{1}{c}{}                                                                                              & \multicolumn{1}{c}{}                                  & \textbf{YouTube}                 & \multicolumn{1}{c}{0.62}                & \multicolumn{1}{c}{0.6}               & \multicolumn{1}{c}{0.64}           & \multicolumn{1}{c}{\textbf{0.66}}  & \multicolumn{1}{c}{0.55}            & \multicolumn{1}{c}{0.51}              & {\ul 0.65}    \\ % \cmidrule{3-10} 
\multicolumn{1}{c}{}                                                                                              & \multicolumn{1}{c}{}                                  & \textbf{X}                       & \multicolumn{1}{c}{0.54}                & \multicolumn{1}{c}{0.59}              & \multicolumn{1}{c}{0.6}            & \multicolumn{1}{c}{NA}             & \multicolumn{1}{c}{0.6}             & \multicolumn{1}{c}{0.58}              & \textbf{0.63} \\ % \cmidrule{3-10} 
\multicolumn{1}{c}{}                                                                                              & \multicolumn{1}{c}{}                                  & \textbf{Reddit}                  & \multicolumn{1}{c}{0.56}                & \multicolumn{1}{c}{\textbf{0.62}}     & \multicolumn{1}{c}{0.61}           & \multicolumn{1}{c}{NA}             & \multicolumn{1}{c}{0.55}            & \multicolumn{1}{c}{0.57}              & \textbf{0.62} \\ \cmidrule{2-10} 
\multicolumn{1}{c}{}                                                                                              & \multicolumn{1}{c}{\multirow{4}{*}{\textbf{YouTube}}} & \textbf{GAB}                     & \multicolumn{1}{c}{0.47}                & \multicolumn{1}{c}{0.52}              & \multicolumn{1}{c}{0.48}           & \multicolumn{1}{c}{\textbf{0.56}}  & \multicolumn{1}{c}{0.5}             & \multicolumn{1}{c}{0.45}              & {\ul 0.53}    \\ % \cmidrule{3-10} 
\multicolumn{1}{c}{}                                                                                              & \multicolumn{1}{c}{}                                  & \textbf{YouTube}                 & \multicolumn{1}{c}{\textbf{0.88}}       & \multicolumn{1}{c}{0.84}              & \multicolumn{1}{c}{{\ul 0.86}}     & \multicolumn{1}{c}{0.79}           & \multicolumn{1}{c}{0.77}            & \multicolumn{1}{c}{0.72}              & 0.76          \\ % \cmidrule{3-10} 
\multicolumn{1}{c}{}                                                                                              & \multicolumn{1}{c}{}                                  & \textbf{X}                       & \multicolumn{1}{c}{0.49}                & \multicolumn{1}{c}{0.53}              & \multicolumn{1}{c}{{\ul 0.57}}     & \multicolumn{1}{c}{NA}             & \multicolumn{1}{c}{0.47}            & \multicolumn{1}{c}{0.45}              & \textbf{0.59} \\ 
% \cmidrule{1-1} % \cmidrule{3-10} 
\multicolumn{1}{c}{\multirow{9}{*}{\textbf{\begin{tabular}[c]{@{}c@{}}Without \\ Target \\ Labels\end{tabular}}}} & \multicolumn{1}{c}{}                                  & \textbf{Reddit}                  & \multicolumn{1}{c}{0.52}                & \multicolumn{1}{c}{0.54}              & \multicolumn{1}{c}{{\ul 0.58}}     & \multicolumn{1}{c}{NA}             & \multicolumn{1}{c}{0.51}            & \multicolumn{1}{c}{0.49}              & \textbf{0.64} \\ \cmidrule{2-10} 
\multicolumn{1}{c}{}                                                                                              & \multicolumn{1}{c}{\multirow{4}{*}{\textbf{X}}}       & \textbf{GAB}                     & \multicolumn{1}{c}{0.62}                & \multicolumn{1}{c}{0.61}              & \multicolumn{1}{c}{0.6}            & \multicolumn{1}{c}{NA}             & \multicolumn{1}{c}{{\ul 0.64}}      & \multicolumn{1}{c}{0.61}              & \textbf{0.65} \\ % \cmidrule{3-10} 
\multicolumn{1}{c}{}                                                                                              & \multicolumn{1}{c}{}                                  & \textbf{YouTube}                 & \multicolumn{1}{c}{0.61}                & \multicolumn{1}{c}{0.62}              & \multicolumn{1}{c}{\textbf{0.65}}  & \multicolumn{1}{c}{NA}             & \multicolumn{1}{c}{0.6}             & \multicolumn{1}{c}{0.57}              & {\ul 0.64}    \\ % \cmidrule{3-10} 
\multicolumn{1}{c}{}                                                                                              & \multicolumn{1}{c}{}                                  & \textbf{X}                       & \multicolumn{1}{c}{0.93}                & \multicolumn{1}{c}{{\ul 0.92}}        & \multicolumn{1}{c}{\textbf{0.93}}  & \multicolumn{1}{c}{NA}             & \multicolumn{1}{c}{0.81}            & \multicolumn{1}{c}{0.78}              & 0.91          \\ % \cmidrule{3-10} 
\multicolumn{1}{c}{}                                                                                              & \multicolumn{1}{c}{}                                  & \textbf{Reddit}                  & \multicolumn{1}{c}{{\ul 0.51}}          & \multicolumn{1}{c}{0.45}              & \multicolumn{1}{c}{0.45}           & \multicolumn{1}{c}{NA}             & \multicolumn{1}{c}{{\ul 0.51}}      & \multicolumn{1}{c}{0.47}              & \textbf{0.54} \\ \cmidrule{2-10} 
\multicolumn{1}{c}{}                                                                                              & \multicolumn{1}{c}{\multirow{4}{*}{\textbf{Reddit}}}  & \textbf{GAB}                     & \multicolumn{1}{c}{0.53}                & \multicolumn{1}{c}{0.57}              & \multicolumn{1}{c}{{\ul 0.63}}     & \multicolumn{1}{c}{NA}             & \multicolumn{1}{c}{0.55}            & \multicolumn{1}{c}{0.56}              & \textbf{0.65} \\ % \cmidrule{3-10} 
\multicolumn{1}{c}{}                                                                                              & \multicolumn{1}{c}{}                                  & \textbf{YouTube}                 & \multicolumn{1}{c}{0.39}                & \multicolumn{1}{c}{0.44}              & \multicolumn{1}{c}{\textbf{0.56}}  & \multicolumn{1}{c}{NA}             & \multicolumn{1}{c}{0.45}            & \multicolumn{1}{c}{0.44}              & {\ul 0.55}    \\ % \cmidrule{3-10} 
\multicolumn{1}{c}{}                                                                                              & \multicolumn{1}{c}{}                                  & \textbf{X}                       & \multicolumn{1}{c}{{\ul 0.55}}          & \multicolumn{1}{c}{0.49}              & \multicolumn{1}{c}{0.54}           & \multicolumn{1}{c}{NA}             & \multicolumn{1}{c}{0.53}            & \multicolumn{1}{c}{0.54}              & \textbf{0.56} \\ % \cmidrule{3-10} 
\multicolumn{1}{c}{}                                                                                              & \multicolumn{1}{c}{}                                  & \textbf{Reddit}                  & \multicolumn{1}{c}{{\ul 0.89}}          & \multicolumn{1}{c}{\textbf{0.9}}      & \multicolumn{1}{c}{{\ul 0.89}}     & \multicolumn{1}{c}{NA}             & \multicolumn{1}{c}{0.79}            & \multicolumn{1}{c}{0.81}              & 0.88          \\ % \hline
\midrule
\multicolumn{3}{c}{\textbf{Average Performance}}                                                                                                                                                              & \multicolumn{1}{c}{0.62}                & \multicolumn{1}{c}{0.63}              & \multicolumn{1}{c}{{\ul 0.65}}     & \multicolumn{1}{c}{0.18}           & \multicolumn{1}{c}{0.6}             & \multicolumn{1}{c}{0.58}              & \textbf{0.66} \\ % \hline
\bottomrule
\end{tabular}%
}
\caption{Cross-platform and in-dataset evaluation results for the different baseline models compared against {\hw}. Boldfaced values denote the best performance, and the underline denotes the second-best performance. NA implies Not Applicable due to absence of target labels.}
\label{tab:results}
\end{table}

\subsection{RQ1. Performance Comparison}
Using macro-F1 for assessment, the generalizability of our model, {\hw} is examined via comparative analysis against several baselines and across multi-platform datasets (Table~\ref{tab:results}). This analysis evaluates the cross-platform performance of {\hw} by emphasizing causal rather than non-causal elements without auxiliary labels. Observations w.r.t RQ.1 are as follows:

\begin{table}[t]
\centering
\footnotesize
\setlength{\tabcolsep}{2pt} % Reduce space between columns
\renewcommand{\arraystretch}{1.2} % Adjust vertical spacing between rows
\begin{tabular}{llcc}
\toprule
\textbf{Source} & \textbf{Model} & \textbf{GAB} & \textbf{YouTube} \\ 
\midrule
\multirow{4}{*}{GAB} & Fully Supervised (CATCH) & \textbf{0.82} & \textbf{0.66}    \\ 
                     & Weak-Supervised-GPT4      & 0.71          & 0.60    \\ 
                     & Unsupervised             & 0.69          & 0.59    \\ 
                     & {\hw}                     & \underline{0.81}    & \underline{0.65}   \\ 
\midrule
\multirow{4}{*}{YouTube} & Fully Supervised (CATCH) & \textbf{0.56}          & \textbf{0.79}    \\ 
                         & Weak-Supervised-GPT4      & 0.51          & \underline{0.78}   \\ 
                         & Unsupervised             & 0.43          & 0.55    \\ 
                         & {\hw}                     & \underline{0.53}    & 0.76    \\ 
\bottomrule
\end{tabular}
\caption{Comparison of {\hw}'s generalization capabilities through macro-F1 against different modeling techniques.}
\label{tab:ablation}
\end{table}

% In examining the generalizability of the weakly supervised model {\hw}, we analyze its performance against several baseline and state-of-the-art models in various real-world datasets. We structure the datasets by source and target platforms and apply macro-F1 as the metric of evaluation, as reflected in Table~\ref{tab:results}. This analysis allows us to investigate the cross-platform efficacy of {\hw}, which aims to disentangle causal from non-causal factors without the aid of auxiliary labels. Our observations regarding the cross-platform performance in relation to RQ.1 are as follows:
\begin{itemize}
    \item The {\hw} model exhibits notable generalizability across different platforms. Despite the absence of auxiliary labels, e.g. the target of hate labels, {\hw} still manages to discern and prioritize causal over non-causal factors. This indicates that the contrastive learning strategy coupled with the confidence based denoising allows {\hw} to maintain performance in diverse environments, reinforcing the model's utility in scenarios where hate targets may not be readily available.
    
    \item When comparing causal models, \textbf{CATCH}, \textbf{PEACE}, and {\hw}, we observe that these models, on average, outperform the non-causal methods, highlighting the importance of causality in model generalization. PEACE performs well even in the absence of target labels, however, PEACE utilizes only two causal cues i.e. sentiment and aggression for learning generalizable representations. However, there may be other factors (e.g. user history, demographics) that are influential in determining hatefulness but not easily quantifiable, limiting PEACE's capabilities. {\hw}, on the other hand, models these cues in the latent space, displaying a competitive edge in settings without target labels, hinting at its efficient use of underlying causal relationships that are consistent across platforms, in contrast to non-causal features that may vary and lead to overfitting.
    
    % \item Within the context of weak supervision, \textbf{OURS}, along with \textbf{XClass} and \textbf{LOTClass}, is tasked with the challenge of maintaining performance without the comprehensive guidance that fully supervised models receive. Despite this, \textbf{OURS} generally exceeds the benchmark set by its weakly supervised peers. This success could be attributed to the effective way in which \textbf{OURS} leverages unlabeled data to capture the essence of hate speech, which is a testament to its sophisticated design in the face of limited supervision.
    
    \item As for models like \textbf{HateBERT} and \textbf{HateXplain}, while powerful and based on robust architectures, they appear to underperform in cross-platform scenarios. This could be indicative of a possible overfitting to platform-specific nuances within the training data, limiting their applicability in a cross-platform context where platform-invariant features are essential. Conversely, {\hw} circumvents this by its very design, which is geared towards recognizing and utilizing features of hate speech that are pertinent to hate and shared across various platform.

    \item Compared to supervised models, both \textbf{XClass} and \textbf{LOTClass} perform close to supervised models, even in the absence of explicit hate labels indicating how self-training and denoising can facilitate qualitative learning for scenarios where labeled data is scarce.
\end{itemize}

\subsection{RQ2. How Helpful Are The Weak-Labels?}
The utilization of weak labels in hate speech detection models serves as an intriguing middle ground between the often resource-intensive fully supervised techniques and the less constrained unsupervised approaches. In this experiment, we investigated the effectiveness of weak labels by contrasting our model—which makes use of these labels—with three different approaches. The first comparison is made with CATCH, a fully supervised technique that allows for a more focused disentanglement by providing both hate and target labels. The second model in our comparative study involves GPT-4, where we crafted a prompt as follows:

\begin{tcolorbox}[
  title= \textbf{Prompt to Detect Hate Targets},
  colback=brown!10,
  colframe=brown!75!black,
  colbacktitle=brown!85!white,
  coltitle=white,
  enhanced,
  sharp corners,
  boxrule=1pt,
  % fonttitle=\sffamily\bfseries\small,
  width=1\columnwidth, % Adjust the width here
  ]
The following examples show the post and the target group being talked about in the post. Examples: ...
Now, given the following posts, identify the main target group of the post. The target category of the post refers to the entity being talked about in the post. The possible categories are Ability/Disability, Class, Gender, Immigration Status, Nationality, Race, Religion, Sexuality, and Sexual Preferences.
\end{tcolorbox}

In order to bridge the gap between fully supervised and unsupervised algorithms for hate speech identification, weak labels are deployed. We evaluated the effectiveness of weak labels using {\hw} in comparison to three approaches: a fully supervised one (CATCH), one that had GPT-4 assistance, and one that was completely unsupervised. As Table~\ref{tab:ablation} shows, GPT-4 performs better than the unsupervised approach, but it falls short of {\hw}.

The training signal may be diluted by noise introduced by the powerful label synthesis of GPT-4, which has a 15-20\% target misidentification rate. In comparison, weak label denoising is essential for reliable feature disentanglement in the presence of noisy data in {\hw}. Furthermore, LLMs such as GPT-4 might not be able to properly recognize the contextually complicated nature of hate speech, while {\hw}'s lax monitoring is more effective.

{\hw} outperforms unsupervised and GPT-4-based approaches in terms of generalization within and between the YouTube and GAB domains. Its ability to transfer hate speech characteristics across platforms with remarkable resilience highlights the efficacy of the weak-label framework. The study shows that improving model generalizability in a variety of contexts can be accomplished at a reasonable cost by using weak labels.

\subsection{RQ3. Are the Disentangled Representations Truly Invariant?}
\begin{figure}[t]
    \centering
    \subfloat[\centering CATCH]{{\includegraphics[width=0.25\textwidth]{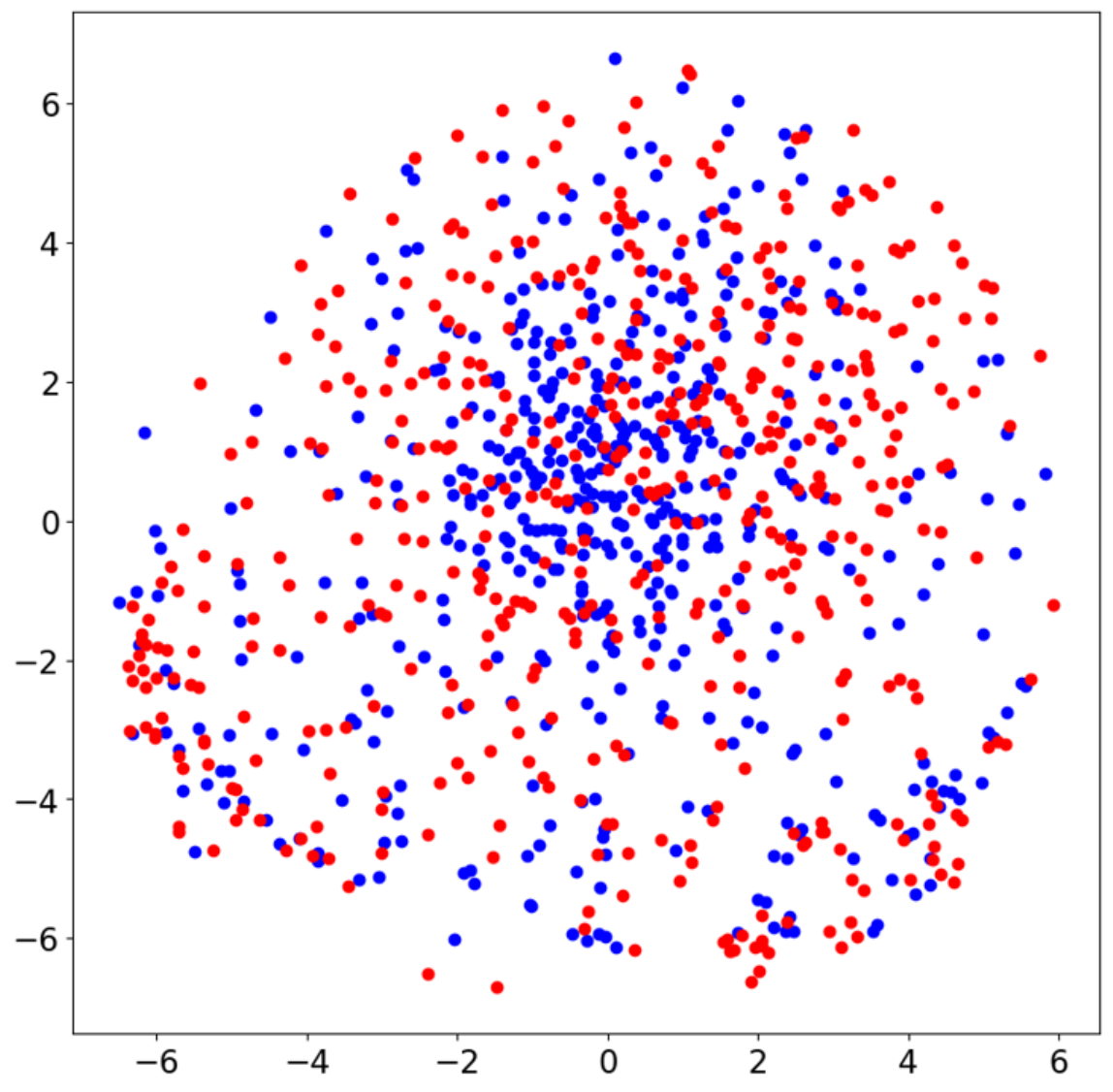}}\label{gab_model}}
    \hfil % Horizontal space between figures
    \subfloat[\centering HateBERT]{{\includegraphics[width=0.25\textwidth]{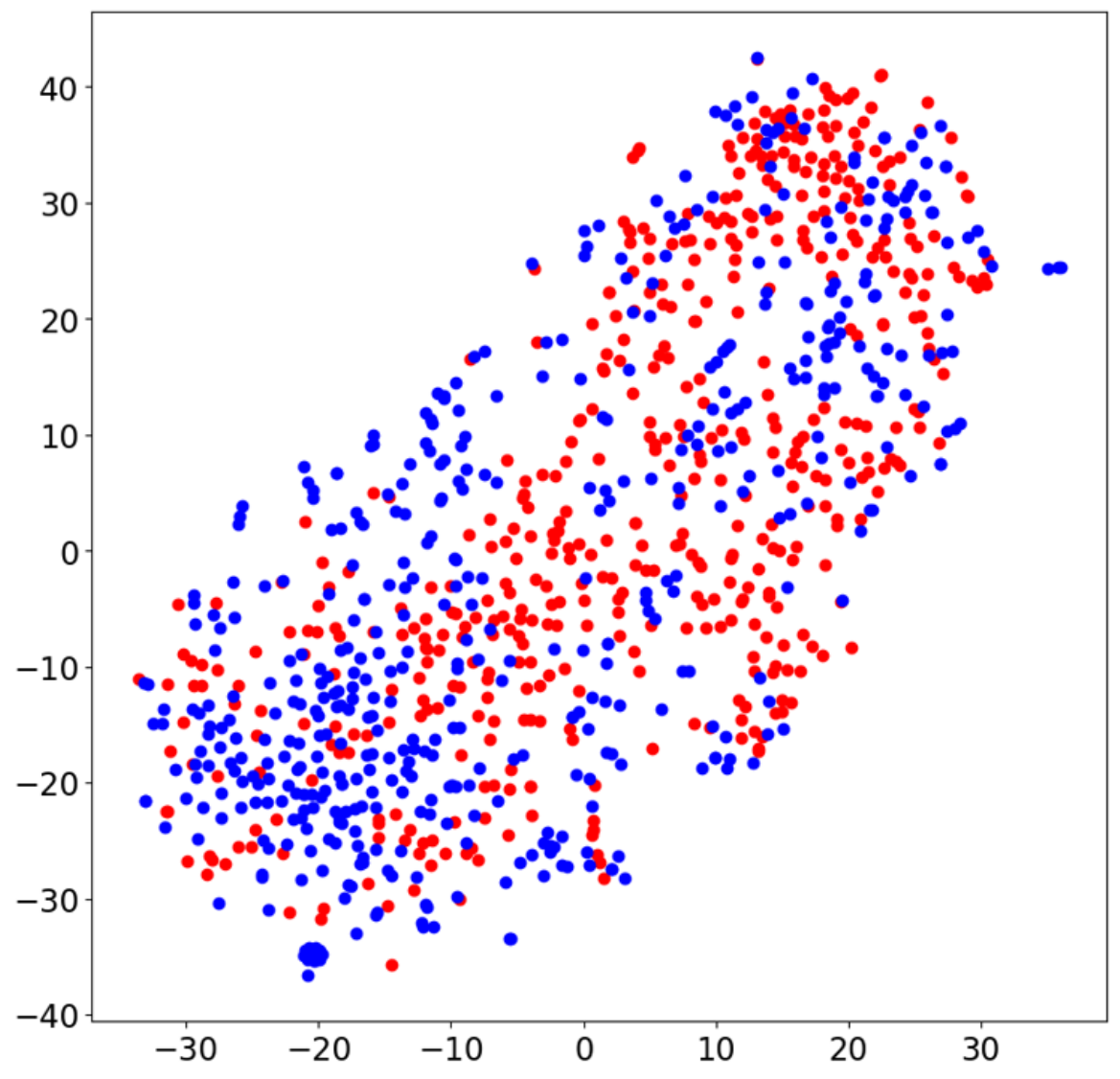}}\label{gab_HB}}
    \hfil % Horizontal space between figures
    \subfloat[\centering PEACE]{{\includegraphics[width=0.25\textwidth]{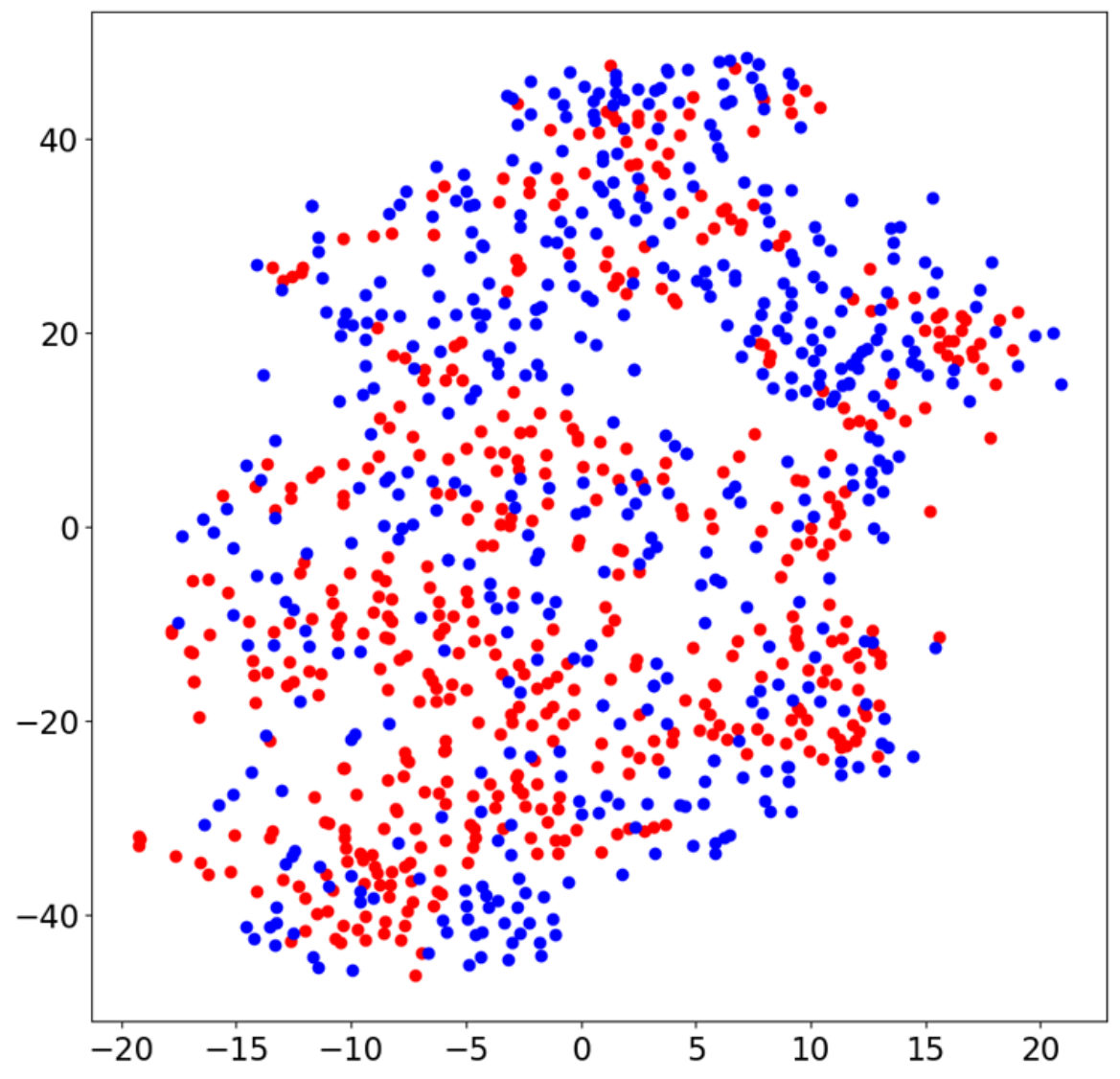}}\label{gab_PEACE}}
    \hfil % Horizontal space between figures
    \subfloat[\centering {\hw}]{{\includegraphics[width=0.25\textwidth]{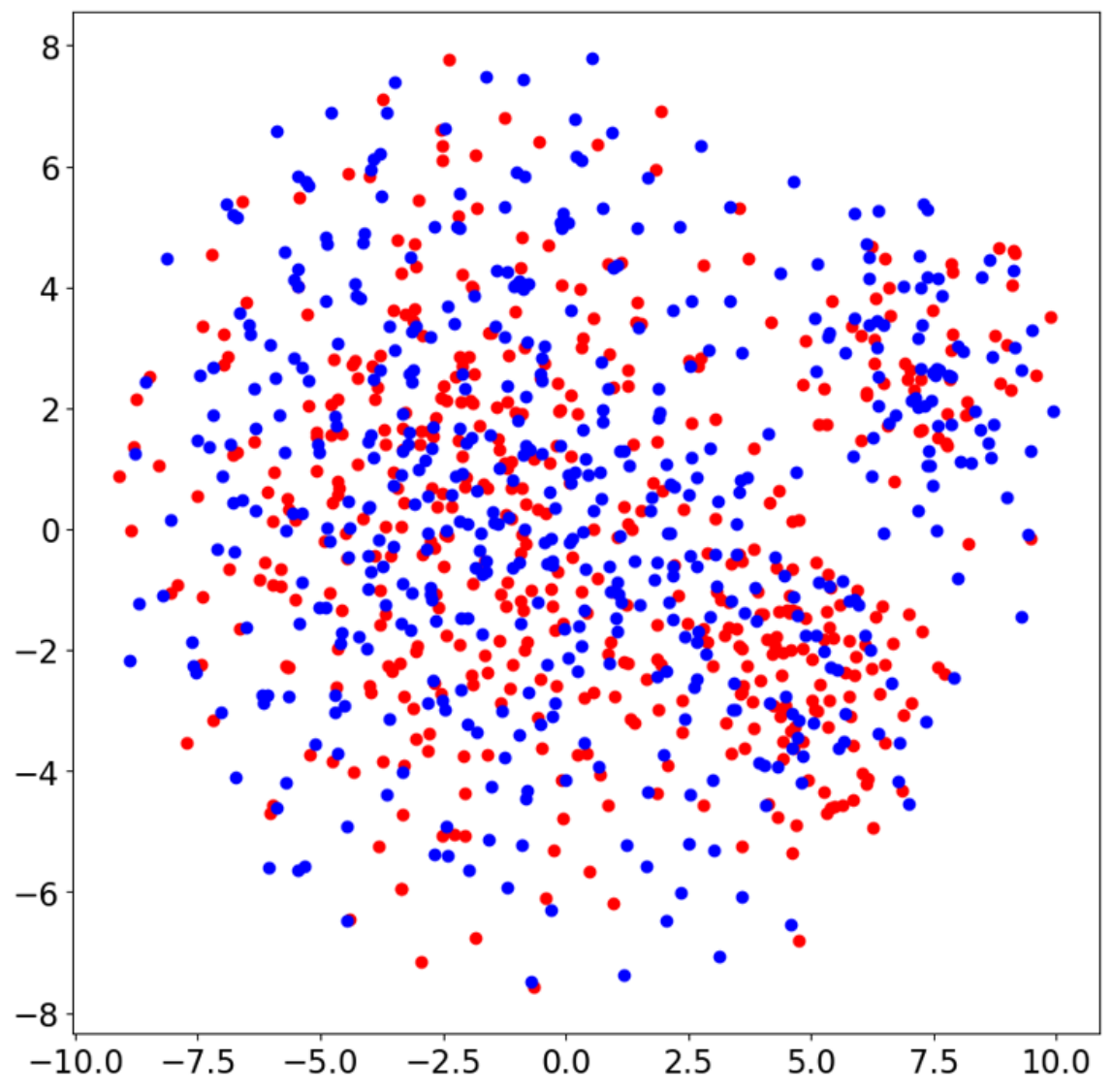}}\label{gab_HE}}
          \\
          \subfloat % The legend, centered alone in the next row
     {\centering\includegraphics[width=0.3\textwidth]{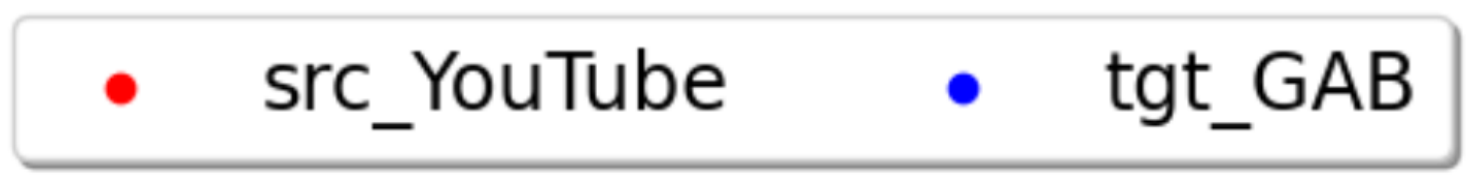}\label{legend}}
     \\
    \caption{Visualizing the representations from different models to verify invariance across platforms. \texttt{src} (\texttt{tgt}) denote the source (target) platforms.}
    
    % \acil{Need to increase the font size of the axes. Also, can we have Hate-Watch be in-line with "(e)"?}
    \label{invar}
\end{figure}

{\hw} focuses on invariant traits that are essential for hate speech identification across platforms, and achieves excellent causal disentanglement without depending on hate target labels. We do an additional experiment to assess the model's ability to learn really invariant features. The experiment's hypothesis is that, as the causal features are common across platforms, there should be significant overlap in the representations of those features if the model can learn invariant features. We trained {\hw} on YouTube and evaluated its generalization on GAB in order to assess this hypothesis. To evaluate representation invariance, we visualized the causal representations (where possible) from 1,000 occurrences per platform, using t-SNE.

{\hw} successfully captures invariant hate speech traits, akin to its fully supervised version, CATCH, according to the t-SNE plots. Among the methods that highlight the advantages of incorporating causality into hate speech detection frameworks are {\hw} and other causal approaches like CATCH and PEACE, which show a notable advantage in learning invariant representations over language-model based systems like HateBERT. We speculate that the reason HateBERT and other similar models can't learn invariantly or generalize well is because they rely too much on platform-specific details (e.g., HateBERT's training data comes from Reddit, thus the model may have picked up on platform peculiarities while learning from the data).

\section{Conclusion and Future Work}
This paper introduced {\hw}, a novel framework that uses weakly supervised causal disentanglement to effectively navigate the challenging terrain of online hate speech detection. {\hw} effectively disentangles platform dependent features from the platform invariant features of hate, enabling robust detection across various platforms. It sets a new benchmark for flexible and scalable content moderation by successfully reducing the need for large amounts of auxiliary labeled data—that is, the target of hate—through confidence-based reweighting and contrastive regularization. Our results highlight {\hw}'s ability to detect hate speech across platforms, leading to the creation of safer online~environments.

% Future works will concentrate on attempting to generalize between various forms of hate, such as explicit to implicit or vice versa, improving its capacity to adjust to the constantly shifting dynamics of online discourse without requiring a great deal of manual labeling. Future versions of {\hw} seek to provide even more powerful tools for countering hate speech by investigating ways to adapt more autonomously to emerging patterns of hate speech. This will ultimately promote more inclusive and respectful digital settings.
Future efforts will aim at bridging the gap between different hate expressions and adapting to the evolving landscape of online discourse with reduced manual intervention making {\hw} effectively identify and adapt to new hate speech patterns, resulting in safer communities.
% will further our goal of fostering safer, more respectful digital communities.
\section{Ethical Statement}
\subsection{Hate Speech Datasets: Usage and Anonymity}
In our research, we have utilized publicly available, well-established datasets, duly citing the relevant sources and adhering to ethical guidelines for their use. We acknowledge the potentially harmful nature of hate speech examples within these datasets, which could be exploited for malicious purposes. Nonetheless, our objective is to enhance the understanding and mitigation of online hate's adverse effects. We have determined that the advantages of employing these real-world examples to explain our research significantly outweigh the associated risks.  
\subsection{Impact Assessment}
The development and deployment of hate speech detection systems necessitate comprehensive impact assessments to gauge their societal implications, concerning freedom of expression and the transparency of detection methods.

\noindent \textbf{Freedom of Expression and Censorship}: Our research is dedicated to creating algorithms capable of identifying and diminishing the presence of harmful language across various platforms, with a keen awareness of the necessity to protect individuals from hate speech while preserving free speech rights. Our methodologies could be applied to content moderation on social media platforms like X, Facebook, and Reddit to filter out hate speech. Nonetheless, an ethical dilemma arises from the possibility of false positives, where non-hateful content might be mistakenly classified as hate speech, potentially infringing upon legitimate free speech. As such, we caution against the sole reliance on our algorithms for real-world content moderation without the complementary judgment of human annotators to make final decisions.

\noindent \textbf{Transparency and Fairness in Detection}: Embracing the values of fairness and impartiality, our work is committed to transparently sharing our methods, findings, and inherent limitations, with a continuous goal of enhancing our system. Our dedication to transparency goes beyond merely disclosing our methodologies and results; it encompasses making our decision-making processes clear and comprehensible to ensure ethical practices are followed.

\bibliographystyle{splncs04}
\bibliography{mybibliography}
\end{document}